\documentclass[journal,web]{article}
\usepackage{PRIMEarxiv}
\usepackage{graphicx}
\usepackage{eucal}
\usepackage{booktabs}
\usepackage{todonotes}
\usepackage{verbatim}
\usepackage{framed,multirow}
\usepackage{amsmath}
\usepackage{amssymb}
\usepackage{xcolor}
\usepackage{color, colortbl}
\usepackage[normalem]{ulem}
\useunder{\uline}{\ul}{}
\usepackage{changepage}
\usepackage{hyperref}
\definecolor{newcolor}{rgb}{.8,.349,.1}
\definecolor{Gray}{gray}{0.9}

\usepackage{todonotes}

\newcommand{\herve}[1]{\todo[fancyline,backgroundcolor=red!25,bordercolor=red]{Herv\'{e} says: #1}}

\begin{document}

\title{Test-Time Adaptation with Shape Moments for Image Segmentation}

\author{Mathilde Bateson, Hervé Lombaert, Ismail Ben Ayed
\\ETS Montréal \\
  \texttt{mathildebateson@gmail.com} \\}


%

\maketitle

\newcommand{\mathbbm}[1]{\text{\usefont{U}{bbm}{m}{n}#1}}

\begin{abstract}
Supervised learning is well-known to fail at generalization under distribution shifts.
In typical clinical settings, the source data is inaccessible and the target distribution is represented with a handful of samples: adaptation 
can only happen at test time on a few (or even a single) subject(s). We investigate test-time \emph{single-subject adaptation} for segmentation, and propose a \emph{Shape-guided Entropy Minimization} objective for tackling this task. During inference for a single testing subject, our loss is minimized with respect to the batch normalization’s scale and bias parameters. We show the potential of integrating various shape priors to guide adaptation to plausible solutions, and validate our method in two challenging scenarios: MRI-to-CT adaptation of cardiac segmentation and cross-site adaptation of prostate segmentation. Our approach exhibits substantially better performances than the existing test-time adaptation methods. Even more surprisingly, it fares better than state-of-the-art domain adaptation methods, although it forgoes training on additional target data during adaptation.
Our results question the usefulness of training on target data in segmentation adaptation, and points to the substantial effect of shape priors on test-time inference.
Our framework can be readily used for integrating various priors and for adapting any segmentation network, and 
our code is available \footnote{\url{https://github.com/mathilde-b/TTA}}.


\end{abstract}

\section{Introduction}

Deep neural networks have achieved state-of-the-art performances in various natural and medical-imaging problems \cite{litjens2017survey}. However, they tend to under-perform when the test-image distribution is different from those seen during training. In medical imaging, this is due to, for instance, variations in imaging modalities and protocols, vendors, machines, clinical sites and subject populations. For semantic segmentation problems, labelling a large number of images for each different target distribution is impractical, time-consuming, and often impossible. To circumvent those impediments, methods learning robust networks with less supervision have triggered interest in medical imaging \cite{Cheplygina2018Notsosupervised}.

This motivates {\em Domain Adaptation} (DA) methods: DA amounts to adapting a model trained on an annotated source domain to another target domain, with no or minimal new annotations for the latter. Popular strategies involve minimizing the discrepancy between source and target distributions in the feature or output spaces \cite{ADDA,tsai2018learning}; integrating a domain-specific module in the network \cite{dou2018pnp}; translating images from one domain to the other \cite{cyclegan}; or integrating a domain-discriminator module and penalizing
its success in the loss function \cite{ADDA}.

In medical applications, separating the source training and adaptation is critical for privacy and regulatory reasons, as the source and target data may come from different clinical sites. Therefore, it is crucial to develop adaptation methods, which neither assume access to the source data nor modify the pre-training stage. Standard DA methods, such as \cite{dou2018pnp,tsai2018learning,ADDA,cyclegan}, do not comply with these restrictions. This has recently motivated \emph{Source-Free Domain Adaptation} (SFDA) \cite{Bateson2020,KARANI2021101907}, a setting where the source data (neither the images nor the ground-truth masks) is unavailable during the training of the adaptation phase. 

Evaluating SFDA methods consists in: (i) adapting on a dedicated training set \textit{Tr} from the target domain; and (ii) measuring the generalization performance on an unseen test set \textit{Te} in the target domain. However, emerging and very recent \emph{Test-Time Adaptation} (TTA) works in machine learning \cite{wang2021tent,Sun2020} and medical imaging \cite{KARANI2021101907,Varsavsky} argue that this is not as useful as adapting directly to the test set \textit{Te}. In various interesting applications, access to the target distribution might not be possible. This is particularly common in medical image segmentation when only a single target-domain subject is available for test-time inference. In the context of image classification, the authors of \cite{wang2021tent} showed recently that simple adaptation of batch normalization’s scale and bias parameters on a set of test-time samples can deal competitively with domain shifts.

With this context in mind, we propose a simple formulation for source-free and single-subject test-time adaptation of segmentation networks. During inference for a single testing subject, we optimize a loss integrating shape priors and the entropy of predictions with respect to the batch normalization’s scale and bias parameters. Unlike the standard SFDA setting, we perform test-time adaptation on each subject separately, and forgo the use of target training set \textit{Tr} during adaptation. Our setting is most similar to the image classification work in \cite{wang2021tent}, which minimized a label-free entropy loss defined over test-time samples. Building on this entropy loss, we further guide segmentation adaptation with domain-invariant shape priors on the target regions, and show the substantial effect of such shape priors on TTA performances.  
We report comprehensive experiments and comparisons with state-of-the-art TTA, SFDA and DA methods, which show the effectiveness of our shape-guided entropy minimization in two different adaptation scenarios: cross-modality cardiac segmentation (from MRI to CT) and prostate segmentation in MRI
across different sites. Our method exhibits substantially better performances than the existing TTA methods. Surprisingly, it also fares better than various state-of-the-art SFDA and DA methods, although it does not train on source and additional target data during adaptation, but just performs joint inference and adaptation on a single 3D data point in the target domain. 
Our results and ablation studies question the usefulness of training on target set \textit{Tr} during adaptation and points to the surprising and substantial effect of embedding shape priors during inference on domain-shifted testing data. 
Our framework can be readily used for integrating various priors and adapting any
segmentation network at test times. 

\begin{figure}[t]
    \includegraphics[width=1\linewidth]{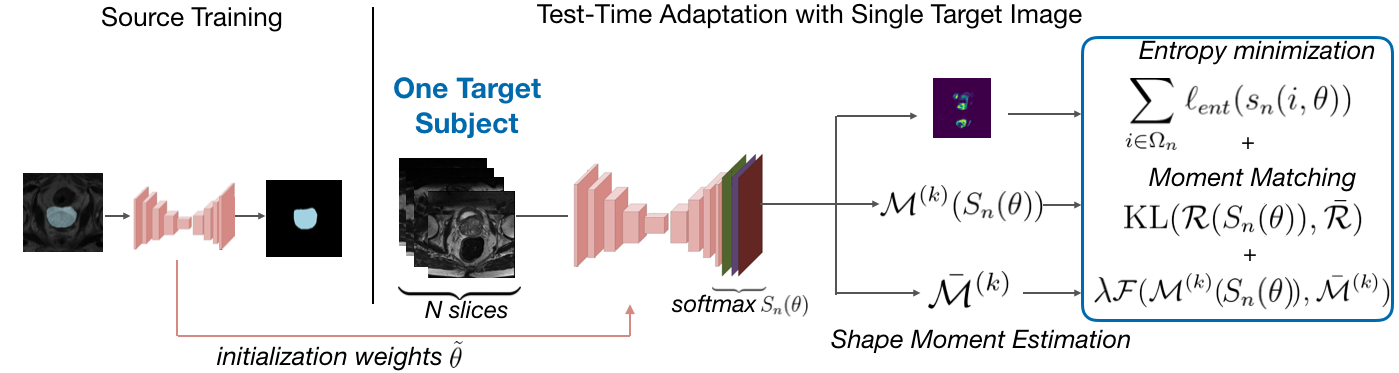}
      \caption[]{Overview of our framework for Test-Time Adaptation with Shape Moments: we leverage entropy minimization and shape priors to adapt a segmentation network on a single subject at test-time.}
      \label{fig:overview}

\end{figure}

\section{Method} 

We consider a set of $M$ source images ${I}_{m}: \Omega_s\subset \mathbb R^{2} \rightarrow {\mathbb R}$, $m=1, \dots, M$, and denote their ground-truth K-class segmentation for each pixel $i \in \Omega_s$ as a $K$-simplex vector ${\mathbf y}_m (i) = \left(y^{(1)}_{m} (i), \dots, y^{(K)}_{m} (i)\right) \in \{0,1\}^K$. For each pixel $i$, its coordinates in the 2D space are represented by the tuple $\left(u_{(i)}, v_{(i)}\right) \in \mathbb{R}^{2}$.

\paragraph{Pre-training Phase} The network is first trained on the source domain only, by minimizing the cross-entropy loss with respect to network parameters $\theta$:  
\begin{equation}\label{eq:crossent}
\begin{aligned}
\min_{\theta} \frac{1}{\left|\Omega_{s}\right|} \sum_{{m}=1}^{M} \ell\left({\mathbf y}_{m} (i), {\mathbf s}_{m} (i, \theta)\right)
  \end{aligned}
\end{equation}
where ${\mathbf s}_{m} (i, \theta) = (s^{(1)}_{m} (i,\theta), \dots, s^{(K)}_{m} (i, \theta)) \in [0,1]^K$ denotes the predicted softmax probability for class $k \in\{1, \ldots, K\}$.
\paragraph{Shape moments and descriptors}
Shape moments are well-known in classical computer vision \cite{nosrati2016incorporating}, and were recently shown useful in the different context of supervised training \cite{KervadecMIDL2021}. Each moment is parametrized by its orders $p, q \in \mathbb{N}$, and each order represents a different characteristic of the shape. For a given $p, q \in \mathbb{N}$ and class $k$, the shape moments of the segmentation prediction of an image $I_n$ can be computed as follows from the softmax matrix $\mathrm{S}_{n}(\theta)=\left(\mathrm{s}_{n}^{(k)}(\theta)\right)_{k=1\ldots K}$ :
$$
\mu_{p, q}\left(\mathrm{s}_{n}^{(k)}(\theta)\right)=\sum_{i \in \Omega} s_{n}^{(k)}(i, \theta) u_{(i)}^{p} v_{(i)}^{q}
$$
Central moments are derived from shape moments to guarantee translation invariance. They are computed as follows:
$$
 \bar{\mu}_{p, q}\left({\mathbf s}_{n}^{(k)}(\theta)\right)=\sum_{i \in \Omega} s_{n}^{k}(i,\theta)\left(u_{(i)}-\bar{u}^{(k)}\right)^{p}\left(v_{(i)}-\bar{v}^{(k)}\right)^{q}.
$$
where 
$\left(\frac{\mu_{1,0}(s_{n}^{(k)}(\theta))}{\mu_{0,0}(s_{n}^{(k)}(\theta))},\frac{\mu_{0,1}(s_{n}^{(k)}(\theta))}{\mu_{0,0}(s_{n}^{(k)}(\theta))}\right)$ are the components of the centroid. 
We use the vectorized form onwards, e.g. $\mu_{p, q}\left(s_{n}(\theta)\right) =\left ( \mu_{p, q}(s_{n}^{(1)}(\theta)), \dots, \mu_{p, q}(s_{n}^{(K)}(\theta)) \right )^\top$.
Building from these definitions, we obtain 2D shape moments from the network predictions. We then derive the shape descriptors $\mathcal{R} ,\mathcal{C},\mathcal{D}$ defined in Table \ref{table:shapes}, which respectively inform on the size, position, and compactness of a shape.  

\vspace{-0.3em}
\begin{table}[h!!!]
\centering
  \caption{Examples of shape descriptors based on softmax predictions.}
\begin{tabular}{ll}
\toprule
Shape Descriptor & \multicolumn{1}{c}{Definition} \\
\midrule
Class-Ratio & $\mathcal{R}(s):=\frac{1}{\left| \Omega_T  \right|}\mu_{0, 0}\left(s\right) $ \\
Centroid & $\mathcal{C}\left(s\right):=\left(\frac{\mu_{1,0}\left(s\right)}{\mu_{0,0}\left(s\right)}, \frac{\mu_{0,1}\left(s\right)}{\mu_{0,0}\left(s\right)}\right)$\\ 
Distance to Centroid & $\mathcal{D}\left(s\right):=\left(\sqrt[2]{\frac{\bar{\mu}_{2,0}\left(s\right)}{\mu_{0,0}\left(s\right)}}, \sqrt[2]{\frac{\bar{\mu}_{0,2}\left(s\right)}{\mu_{0,0}\left(s\right)}}\right) $\\ 
\bottomrule
\label{table:shapes}
\end{tabular}
\end{table}

\vspace{-1.5em}
\paragraph{Test-time adaptation and inference with shape-prior constraints}

Given a single new subject in the target domain composed of $N$ 2D slices, ${I}_n: \Omega_t\subset \mathbb R^{2} \rightarrow {\mathbb R}$, $n=1, \ldots, N$, the first loss term in our adaptation phase is derived from \cite{wang2021tent}, to encourage high confidence in the softmax predictions, by minimizing their weighted Shannon entropy: $\ell_{ent}({\mathbf s}_n (i,\theta)) = - \sum_k \nu_k s^k_n (i,\theta) \log s^k_n (i, \theta)$, where $\nu_k, k=1 \ldots K$, are class weights added to mitigate imbalanced class-ratios.

Ideally, to guide adaptation, for each slice $I_n$, we would penalize the deviations between the shape descriptors of the softmax predictions ${S}_{n}(\theta)$ and those corresponding to the ground truth $\mathbf{y_n}$. As the ground-truth labels are unavailable, instead, we estimate the shape descriptors using the predictions from the whole subject $  \left\{  S_{n}(\theta), n=1,\dots,N\right\}$, which we denote respectively $\mathcal{\bar{C}},\mathcal{\bar{D}}$.

The first shape moment we leverage is the simplest: a zero-order class-ratio $\mathcal{R}$. Seeing these class ratios as distributions, we integrate a KL divergence with the Shannon entropy: 
\begin{equation}\label{eq:AdaMI}
\begin{aligned}
 \mathcal{L}_{TTAS}(\theta) = \sum_n\frac{1}{\left|\Omega_{n}\right|} \sum_{i \in \Omega_t} \ell_{ent}({s}_n (i, \theta))+ \mbox{KL}(\mathcal{R}(S_n (\theta)),\mathcal{\bar{R}}).
  \end{aligned}
\end{equation}
It is worth noting that, unlike \cite{Bateson2022}, which used a loss of the form in Eq~\eqref{eq:AdaMI} for training on target data, here we use this term for inference on a test subject, as a part of our overall shape-based objective.
Additionally, we integrate the centroid ($\mathcal{M}=\mathcal{C}$) and the distance to centroid ($\mathcal{M}=\mathcal{D}$) to further guide adaptation to plausible solutions:
\begin{equation}
\begin{aligned}
\label{eq:ineq}
   \min_{\theta} &\quad\mathcal{L}_{TTAS}(\theta) 
   \\
  &\text{s.t. } \left|\mathcal{M}^{(k)}(S_n (\theta))-\mathcal{\bar{M}}^{(k)} \right|\leq0.1, & k=\{2,\dots, K\}, n = \{1, \dots, N\}.
  \end{aligned}
\end{equation}


Imposing such hard constraints is typically handled through the minimization of the Lagrangian dual in standard convex-optimization. As this is computationally intractable in deep networks, inequality constraints such as Eq~\eqref{eq:ineq} are typically relaxed to soft penalties \cite{He2017,kervadec2019constrained,Jia2017}.
Therefore, we experiment with the integration of $\mathcal{C}$ and $\mathcal{D}$ through a quadratic penalty, leading to the following unconstrained objectives for joint test-time adaptation and inference:
\begin{equation}\label{eq:TTAS}
 \sum_n\frac{1}{\left|\Omega_{t}\right|} \sum_{i \in \Omega_n} \ell_{ent}({\mathbf s}_n (i, \theta))+ \mbox{KL}(\mathcal{R}(S_n (\theta)),\mathcal{\bar{R}}) + \lambda \mathcal{F}(\mathcal{M}(S_n (\theta)),\mathcal{\bar{M}}),
\end{equation}
where $\mathcal{F}$ is a quadratic penalty function corresponding to the relaxation of Eq~\eqref{eq:ineq}: $\mathcal{F}(m_1,m_2)= [m_1-0.9m_2]_+^2 + [1.1m_2-m_1]_+^2$ and $[m]_+ = \max (0,m)$, with $\lambda$ denoting a weighting hyper-parameter.
Following recent TTA methods \cite{wang2021tent,KARANI2021101907}, we only optimize for the scale and bias parameters of batch normalization layers while the rest of the network is frozen. Figure \ref{fig:overview} shows the overview of the proposed framework.

\section{Experiments}
    
\subsection{Test-time Adaptation with shape descriptors} 

\paragraph{\textbf{Heart Application}} We employ the 2017 Multi-Modality Whole Heart Segmentation (MMWHS) Challenge dataset for cardiac segmentation \cite{Zhuang2019}. The dataset consists of 20 MRI (source domain) and 20 CT volumes (target domain) of non-overlapping subjects, with their manual annotations of four cardiac structures: the Ascending Aorta (AA), the Left Atrium (LA), the Left Ventricle (LV) and the Myocardium (MYO). We employ the pre-processed data provided by \cite{dou2018pnp}. 
The scans were normalized as zero mean and unit variance, and data augmentation based on affine transformations was performed. For the domain adaptation benchmark methods (DA and SFDA), we use the data split in \cite{dou2018pnp}: 14 subjects for training, 2 for validation, and 4 for testing. Each subject has $N=256$ slices.

\paragraph{\textbf{Prostate Application}} We employ the dataset from the publicly available NCI-ISBI 2013 Challenge\footnote{https://wiki.cancerimagingarchive.net}. It is composed of manually annotated T2-weighted MRI from two different sites: 30 samples from Boston Medical Center (source domain), and 30 samples from Radboud University Medical Center (target domain). For the DA and SFDA benchmark methods, 19 scans were used for training, one for validation, and 10 scans for testing. We used the pre-processed dataset from \cite{SAML}, who resized each sample to $384\times384$ in axial plane, and normalized it to zero mean and unit variance. We employed data augmentation based on affine transformations on the source domain. Each subject has $N \in \left [15,24 \right ] $ slices.

\paragraph{\textbf{Benchmark Methods}} Our first model denoted $TTAS_{\mathcal{R}\mathcal{C}}$ constrains the class-ratio $\mathcal{R}$ and the centroid $\mathcal{C}$ using Eq~\eqref{eq:TTAS}; similarly, $TTAS_{\mathcal{R}\mathcal{D}}$ constrains $\mathcal{R}$ and the distance-to-centroid $\mathcal{D}$. We compare to two \emph{TTA} methods: the method in \cite{KARANI2021101907}, denoted $TTDAE$, where an auxiliary branch is used to denoise segmentation, and $Tent$ \cite{wang2021tent}, which is based on the following loss: $\min_{\theta}\sum_{n} \sum_{i \in \Omega_n} \ell_{ent}({\mathbf s}_n (i, \theta))$. Note that $Tent$ corresponds to performing an ablation of both shape moments terms in our loss. As an additional ablation study, $TTAS_{\mathcal{R}}$ is trained with the class-ratio matching loss in Eq~\eqref{eq:AdaMI} only.
We also compared to two \emph{DA} methods based on class-ratio matching, \textit{CDA} \cite{Bateson2021}, and $CurDA$ \cite{zhang2019curriculum}, and to the recent source-free domain adaptation (\emph{SFDA}) method $AdaMI$ in \cite{Bateson2022}. 
A model trained on the source only, \textit{NoAdap}, was used as a lower bound. A model trained on the target domain with the cross-entropy loss, $Oracle$, served as an upper bound.

\paragraph{\textbf{Estimating the shape descriptors}}\label{sec:sizeprior} For the estimation of the class-ratio $\mathcal{\bar{R}}$, we employed the coarse estimation in \cite{Bateson2021}, which is derived from anatomical knowledge available in the clinical literature.
For $\mathcal{M} \in \left\{ \mathcal{C},\mathcal{D}\right\}$, we estimate the target shape descriptor from the network prediction masks $\mathbf{\hat{y}_n}$ after each epoch: $\mathcal{\bar{M}}^{(k)} = \frac{1}{\left|V^k \right|}\sum_{v \in V^{k}}v$, with  $V^{k} = \left\{ {\mathcal{M}^{(k)}(\mathbf{\hat{y}_n}) \text{ if }{\mathcal{R}^k(\mathbf{\hat{y}_n})>\epsilon^k}},n=1 \cdots N \right\}$.

Note that, for a fair comparison, we used exactly the same class-ratio priors and weak supervision employed in the benchmarks methods in \cite{Bateson2021,Bateson2022,zhang2019curriculum}.
Weak supervision takes the form of simple image-level tags by setting $\mathcal{\bar{R}}^{(k)}=\textbf{0}$ and $\lambda=0$ for the target images that do not contain structure $k$.   

\paragraph{\textbf{Training and implementation details}} For all methods, the segmentation network employed was UNet \cite{UNet}. A model trained on the source data with Eq~\eqref{eq:crossent} for 150 epochs was used as initialization. Then, for TTA models, adaptation is performed on each test subject independently, without target training. Our model was initialized with Eq~\eqref{eq:AdaMI} for 150 epochs, after which the additional shape constraint was added using Eq~\eqref{eq:TTAS} for 200 epochs.  As there is no learning and validation set in the target domain, the hyper-parameters are set following those in the source training, and are fixed across experiments: we trained with the Adam optimizer \cite{Adam}, a batch size of $min(N,22)$, an initial learning rate of $5\times10^{-4}$, a learning rate decay of 0.9 every 20 epochs, and a weight decay of $10^{-4}$. The weights $\nu_k$ are calculated as: $\nu_k=\frac{\bar{\mathcal{R}}_k^{-1}}{\sum_k \bar{\mathcal{R}}_k^{-1}}$. We set $\lambda=1\times10^{-4}$.

\paragraph{\textbf{Evaluation}} The 3D Dice similarity coefficient (DSC) and the 3D Average Surface distance (ASD) were used as evaluation metrics in our experiments.

\subsection{Results \& discussion} 


Table~\ref{table:resultswhs} and Table \ref{table:resultspro} report quantitative metrics for the heart and prostate respectively. Among DA methods, the source-free $AdaMI$ achieves the best DSC improvement over the lower baseline \textit{NoAdap}, with a mean DSC of 75.7\% (cardiac) and 79.5\% (prostate). Surprisingly though, in both applications, our method $TTAS_{\mathcal{R}\mathcal{D}}$ yields better scores: 76.5\% DSC, 5.4 vox. ASD (cardiac) and 79.5\% DSC, 3.9 vox. ASD (prostate); while $TTAS_{\mathcal{R}\mathcal{C}}$ achieves the best DSC across methods: 80.0\% DSC and 5.3 vox. ASD (cardiac), 80.2\% DSC and 3.79 ASD vox. (prostate).
Finally, comparing to the TTA methods, both $TTAS_{\mathcal{R}\mathcal{C}}$ and $TTAS_{\mathcal{R}\mathcal{D}}$ widely outperform $TTADAE$, which yields 40.7\% DSC, 12.9 vox. ASD (cardiac) and 73.2\% DSC, 5.80 vox. ASD (prostate), and $Tent$, which reaches 48.2\% DSC, 11.2 vox. ASD (cardiac) and 68.7\% DSC, 5.87 vox. ASD (prostate). 

\begin{table}[t]
\footnotesize
    \caption{Test-time metrics on the cardiac dataset, for our method and various \textit{Domain Adaptation} (DA), \textit{Source Free Domain Adaptation} (SFDA) and \textit{Test Time Adaptation} (TTA) methods.}
   \centering
    \resizebox{\textwidth}{!}{
      \begin{tabular}{lccccccc|cp{5pt}cccc|c}

\toprule

\multirow{2}{3em}{Methods} & \multirow{2}{3em}{DA}& \multirow{2}{3em}{SFDA} & \multirow{2}{3em}{TTA} &  
 \multicolumn{5}{c}{ DSC (\%)} && \multicolumn{5}{c}{ASD (vox)} \\
 \cmidrule(lr){5-9}\cmidrule(lr){10-15}
& &  & & AA & LA & LV & Myo & Mean && AA & LA & LV & Myo & Mean
\\
\hline
NoAdap (lower b.)&&&& 49.8&62.0&21.1&22.1&38.8 && 19.8&13.0&13.3&12.4&14.6  \\
Oracle \, \,(upper b.)&&&&91.9  & 88.3  & 91.0  & 85.8  & 89.2 && 3.1 & 3.4 & 3.6 & 2.2 & 3.0 \\
\midrule
CurDA \cite{zhang2019curriculum}&$\checkmark$&$\times$& $\times$&79.0  & 77.9   & 64.4 & 61.3 & 70.7&& 6.5 & 7.6 & 7.2 & 9.1 & 7.6 \\
CDA \cite{Bateson2021} & $\checkmark$& $\times$&$\times$&77.3 & 72.8 & 73.7 & 61.9 &71.4 && \textbf{4.1} &6.3 &  6.6 &6.6 & 5.9 \\
AdaMI \cite{Bateson2022} & $\times$&$\checkmark$&$\times$ &83.1&78.2&74.5&66.8& 75.7&&5.6&\textbf{4.2}&\textbf{5.7}&6.9&5.6\\
TTDAE \cite{KARANI2021101907} &$\times$& $\times$& $\checkmark$& 59.8 & 26.4 & 32.3& 44.4 & 40.7 && 15.1 & 11.7 & 13.6 &11.3 & 12.9 \\
Tent \cite{wang2021tent} & $\times$& $\times$& $\checkmark$& 55.4 & 33.4 &63.0 &41.1 & 48.2 && 18.0 & 8.7 & 8.1 & 10.1 & 11.2 \\
\rowcolor{Gray}\begin{tabular}[c]{@{}l@{}}Proposed Method \end{tabular}  &&  &  & & & & &&  & & & & &\\
\textbf{TTAS$_{\mathcal{RC}}$} (Ours)  &$\times$&$\times$&$\checkmark$&\textbf{85.1}&\textbf{82.6}&\textbf{79.3}&\textbf{73.2}& \textbf{80.0}&&5.6&4.3&6.1&\textbf{5.3}&\textbf{5.3}\\
\textbf{TTAS$_{\mathcal{RD}}$} (Ours) &$\times$&$\times$&$\checkmark$&82.3&78.9&76.1&68.4& 76.5&&4.0&5.8&6.1&5.7&5.4\\
\rowcolor{Gray}\begin{tabular}[c]{@{}l@{}}Ablation study \end{tabular}&  &   &  & & & & &&  & & & & &\\
\textbf{TTAS$_{\mathcal{R}}$}  &$\times$&$\times$&$\checkmark$ &78.9&77.7&74.8&65.3& 74.2&& 5.2&4.9&7.0&7.6&6.2\\
\bottomrule
  \end{tabular}
  }
  \label{table:resultswhs}
\end{table}

\vspace{-0.5em}
\begin{table}[h!]
\centering
\footnotesize
\caption{Test-time metrics on the prostate dataset.}

\begin{tabular}{lccccc}
\toprule
Methods & DA & SFDA &TTA & DSC (\%) & ASD (vox) \\

\midrule
NoAdap (lower bound) &&&    & 67.2 & 10.60\\
Oracle \, \,(upper bound)  &&&   & 88.9 & 1.88\\
\midrule 
\begin{tabular}[c]{@{}l@{}}CurDA \cite{zhang2019curriculum}\end{tabular} &  $\checkmark$ &$\times$& $\times$  & 76.3 & 3.93\\
\begin{tabular}[c]{@{}l@{}}CDA \cite{Bateson2021}\end{tabular} &  $\checkmark$&$\times$  &$\times$  & 77.9 & \textbf{3.28}\\
\begin{tabular}[c]{@{}l@{}}AdaMI\cite{Bateson2022}\end{tabular} & $\times$&$\checkmark$& $\times$& 79.5 & 3.92\\
\begin{tabular}[c]{@{}l@{}}TTDAE \cite{KARANI2021101907}\end{tabular} &$\times$ & $\times$ & $\checkmark$ & 73.2 & 5.80\\
\begin{tabular}[c]{@{}l@{}}Tent \cite{wang2021tent}\end{tabular} &$\times$  &$\times$ & $\checkmark$ & 68.7 & 5.87\\
\rowcolor{Gray}\begin{tabular}[c]{@{}l@{}}Proposed Method \end{tabular}  & & & & & \\
\begin{tabular}[c]{@{}l@{}}TTAS$_{\mathcal{RC}}$ (Ours)\end{tabular} &  $\times$& $\times$ & $\checkmark$& \textbf{80.2} & 3.79\\
\begin{tabular}[c]{@{}l@{}}TTAS$_{\mathcal{RD}}$ (Ours)\end{tabular} &  $\times$& $\times$& $\checkmark$ & 79.5 & 3.90\\
\rowcolor{Gray}\begin{tabular}[c]{@{}l@{}}Ablation study \end{tabular} & &  & & & \\
\begin{tabular}[c]{@{}l@{}}TTAS$_{\mathcal{R}}$ (Ours)\end{tabular} &  $\times$& $\times$& $\checkmark$ & 75.3 & 5.06\\
\bottomrule
\end{tabular}
\label{table:resultspro}
\end{table}

\begin{figure}[t]
\centering
    \includegraphics[width=\textwidth]{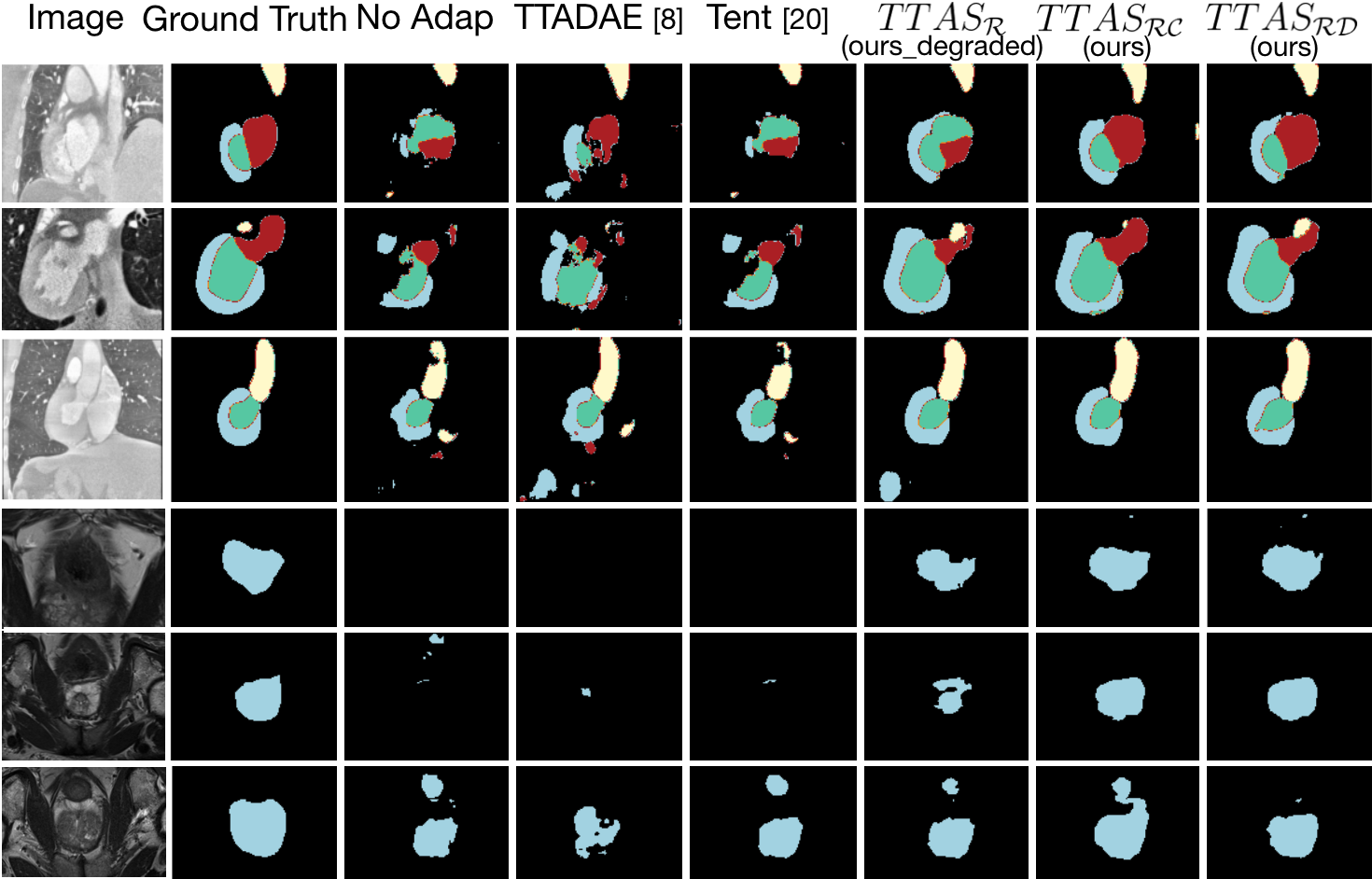}
      \caption[]{Qualitative performance on cardiac images (top) and prostate images (bottom): examples of the segmentations achieved by our formulation ($TTAS_{\mathcal{R}\mathcal{C}},TTAS_{\mathcal{R}\mathcal{D}}$), and benchmark TTA models. The cardiac structures of MYO, LA, LV and AA are depicted in blue, red, green and yellow respectively. }
         \label{fig:seg}
\end{figure}

Qualitative segmentations are depicted in Figure~\ref{fig:seg}. These visuals results confirm that without adaptation, a model trained only on source data cannot properly segment the structures on the target images. The segmentation masks obtained using the TTA formulations $Tent$ \cite{wang2021tent}, $TTADAE$ \cite{KARANI2021101907} only show little improvement. Both methods are unable to recover existing structures when the initialization $NoAdap$ fails to detect them (see fourth and fifth row, Figure~\ref{fig:seg}). On the contrary, those produced from our degraded model $TTAS_\mathcal{R}$ show more regular edges and is closer to the ground truth. However, the improvement over $TTAS_\mathcal{R}$ obtained by our two models $TTAS_{\mathcal{R}\mathcal{C}}$, $TTAS_{\mathcal{R}\mathcal{D}}$ is remarkable regarding the shape and position of each structures: the prediction masks show better centroid position (first row, Figure~\ref{fig:seg}, see LA and LV) and better 
compactness (third, fourth, fifth row, Figure~\ref{fig:seg}).

\section{Conclusion}

In this paper, we proposed a simple formulation for \emph{single-subject} test-time adaptation (TTA), which does not need access to the source data, nor the availability of a target training data. 
Our approach performs inference on a test subject by minimizing the entropy of predictions and a class-ratio prior over batchnorm parameters. To further guide adaptation, we integrate shape priors through penalty constraints. 
We validate our method on two challenging tasks, the MRI-to-CT adaptation of cardiac segmentation and the cross-site adaptation of prostate segmentation. Our formulation achieved better performances than state-of-the-art TTA methods, with a 31.8\% (resp. 7.0\%) DSC improvement on cardiac and prostate images respectively. Surprisingly, it also fares better than various state-of-the-art DA and SFDA methods. These results highlight the effectiveness of shape priors on test-time inference, and question the usefulness of training on target data in segmentation adaptation. Future work will involve the introduction of higher-order shape moments, as well as the integration of multiple shapes moments in the adaptation loss. Our test-time adaptation framework is straightforward to use with any segmentation network architecture.





\bibliographystyle{splncs04}
\bibliography{biblio}


\end{document}